# Learning Graphical Models of Images, Videos and Their Spatial Transformations

**Brendan J. Frey**
Computer Science
University of Waterloo
http://www.cs.uwaterloo.ca/~frey

**Nebojsa Jojic**
Electrical and Computer Engineering
University of Illinois at Urbana
http://www.ifp.uiuc.edu/~jojic

## Abstract

Mixtures of Gaussians, factor analyzers (probabilistic PCA) and hidden Markov models are staples of static and dynamic data modeling and image and video modeling in particular. We show how topographic transformations in the input, such as translation and shearing in images, can be accounted for in these models by including a discrete transformation variable. The resulting models perform clustering, dimensionality reduction and time-series analysis in a way that is invariant to transformations in the input. Using the EM algorithm, these transformation-invariant models can be fit to static data and time series. We give results on filtering microscopy images, face and facial pose clustering, handwritten digit modeling and recognition, video clustering, object tracking, and removal of distractions from video sequences.

## 1 INTRODUCTION

Graphical models and their dynamic variants are growing in popularity as a way to build appearance-based models for computer vision.[1] However, many appearance-based models – such as mixtures of Gaussians, factor analyzers, mixtures of factor analyzers and hidden Markov models with Gaussian outputs – are extremely sensitive to spatial or topographic transformations of the input images. These transformations may include translation, rotation, shearing and warping. For example, if a mixture of Gaussians is fit to a set of images that include random translations, the centers will represent different transformations of essentially the same data. Taking face images as an example, it would be more useful for the different clusters to represent different poses and expressions, instead of very noisy versions of different translations.

Imagine what happens to the point in the $N$-dimensional space corresponding to an $N$-pixel image of an object, while the object is deformed by shearing. A very small amount of shearing will move the point only slightly, so deforming the object by shearing will trace a continuous curve in the space of pixel intensities. As illustrated in Fig. 1a, extensive levels of shearing will produce a highly nonlinear curve (consider shearing a thin vertical line), although the curve can be approximated by a straight line locally.

Linear approximations of the transformation manifold have been used to significantly improve the performance of feedforward discriminative classifiers such as nearest neighbors and multilayer perceptrons (Simard et al., 1993). Linear generative models (factor analyzers, mixtures of factor analyzers) have also been modified using linear approximations of the transformation manifold to build in some degree of transformation invariance (Hinton et al., 1997).

In general, the linear approximation is accurate for transformations that couple neighboring pixels, but is inaccurate for transformations that couple nonneighboring pixels. In some applications (e.g., handwritten digit recognition), the input can be blurred so that the linear approximation becomes more robust.

For significant levels of transformation, the nonlinear manifold can be better modeled using a discrete approximation. For example, the curve in Fig. 1a can be represented by a set of points (filled discs). In this approach, a discrete set of possible transformations is specified beforehand and parameters are learned so that the model is invariant to the set of transformations. This approach has been used to design "convolutional neural networks" that are invariant to translation (Le Cun et al., 1998).

We describe how invariance to a discrete set of *known* transformations (like translation) can be built into a generative graphical model and we show how an EM algorithm for the original density model can be extended to the new model by computing expectations over the set of transformations. In earlier work, we illustrated this approach for mixtures of Gaussians (Frey and Jojic, 1999a), factor analyzers (Frey and Jojic, 1999b) and for mixtures of factor analyzers (Jojic and Frey 1999). In this paper, we review this approach and then extend it to dynamic graphical models. The time needed for exact inference and learning in these models scales exponentially with the number of dimensions in the transformation manifold. However, we give results for 7 different types of experiment involving translation and shearing in images and video sequences and find that the algorithm is reasonably fast (it learns in minutes or hours) and very effective at transformation-invariant density modeling.

---

[1] The term "appearance-based model" refers to a model of image and video pixel intensities.

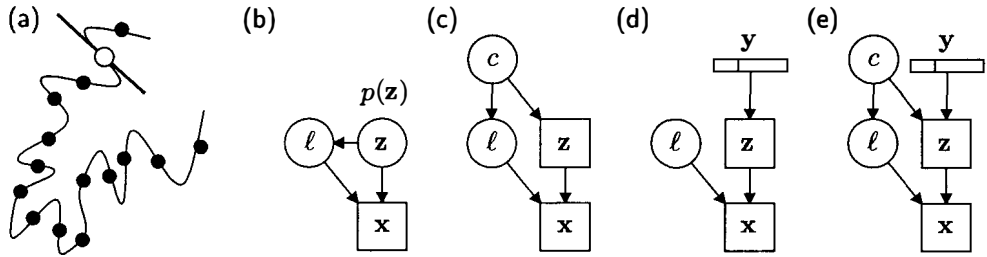

Figure 1: (a) An $N$-pixel greyscale image is represented by a point (unfilled disc) in an $N$-dimensional space. When the object being imaged is deformed by shearing, the point moves along a continuous curve. Locally, the curve is linear, but high levels of shearing produce a highly nonlinear curve, which we approximate by discrete points (filled discs) indexed by $\ell$. (b) A graphical model showing how a discrete transformation variable $\ell$ can be added to a density model $p(\mathbf{z})$ for a *latent image* $\mathbf{z}$ to model the observed image $\mathbf{x}$. The Gaussian pdf $p(\mathbf{x}|\ell, \mathbf{z})$ captures the $\ell$th transformation plus a small amount of pixel noise. (We use a box to represent variables that have Gaussian conditional pdfs.) We have explored (c) transformed mixtures of Gaussians, where $c$ is a discrete cluster index; (d) transformed component analysis (TCA), where $\mathbf{y}$ is a vector of Gaussian factors, some of which may model locally linear transformation perturbations; and (e) mixtures of transformed component analyzers, or transformed mixtures of factor analyzers.

## 2 TRANSFORMATION AS A DISCRETE HIDDEN VARIABLE

We represent transformation $\ell$ by a sparse transformation generating matrix $\mathbf{G}_\ell$ that operates on a vector of pixel intensities. For example, integer-pixel translations of an image can be represented by permutation matrices. Although other types of transformation matrix may not be accurately represented by permutation matrices, many useful types of transformation can be represented by sparse transformation matrices. For example, rotation and blurring can be represented by matrices that have a small number of nonzero elements per row (*e.g.*, at most 6 for rotations).

The observed image $\mathbf{x}$ is linked to the nontransformed *latent image* $\mathbf{z}$ and the transformation index $\ell \in \{1, \ldots, L\}$ as follows:

$$p(\mathbf{x}|\ell, \mathbf{z}) = \mathcal{N}(\mathbf{x}; \mathbf{G}_\ell \mathbf{z}, \mathbf{\Psi}), \qquad (1)$$

where $\mathbf{\Psi}$ is a diagonal matrix of pixel noise variances.

Since the probability of a transformation may depend on the latent image, the joint distribution over the latent image $\mathbf{z}$, the transformation index $\ell$ and the observed image $\mathbf{x}$ is

$$p(\mathbf{x}, \ell, \mathbf{z}) = \mathcal{N}(\mathbf{x}; \mathbf{G}_\ell \mathbf{z}, \mathbf{\Psi}) P(\ell|\mathbf{z}) p(\mathbf{z}). \qquad (2)$$

The corresponding graphical model is shown in Fig. 1b. For example, to model noisy transformed images of just one shape, we choose $p(\mathbf{z})$ to be a Gaussian distribution.

### 2.1 TRANSFORMED MIXTURES OF GAUSSIANS (TMG)

Fig. 1c shows the graphical model for a TMG (Frey and Jojic 1999a), where different clusters may have different transformation probabilities. Cluster $c$ has mixing proportion $\pi_c$, mean $\boldsymbol{\mu}_c$ and diagonal covariance matrix $\boldsymbol{\Phi}_c$. The joint distribution is

$$p(\mathbf{x}, \ell, \mathbf{z}, c) = \mathcal{N}(\mathbf{x}; \mathbf{G}_\ell \mathbf{z}, \mathbf{\Psi}) \mathcal{N}(\mathbf{z}; \boldsymbol{\mu}_c, \boldsymbol{\Phi}_c) \rho_{\ell c} \pi_c, \quad (3)$$

where the probability of transformation $\ell$ for cluster $c$ is $\rho_{\ell c}$. Marginalizing over the latent image gives the cluster/transformation conditional likelihood,

$$p(\mathbf{x}|\ell, c) = \mathcal{N}(\mathbf{x}; \mathbf{G}_\ell \boldsymbol{\mu}_c, \mathbf{G}_\ell \boldsymbol{\Phi}_c \mathbf{G}_\ell^T + \mathbf{\Psi}), \qquad (4)$$

which can be used to compute $p(\mathbf{x})$ and the cluster/transformation responsibility $P(\ell, c|\mathbf{x})$.

This likelihood looks like the likelihood for a mixture of factor analyzers. However, whereas the likelihood computation for $N$ latent pixels takes order $N^3$ time in a mixture of factor analyzers, it takes *linear* time, order $N$, in a TMG, because $\mathbf{G}_\ell \boldsymbol{\Phi}_c \mathbf{G}_\ell^T + \mathbf{\Psi}$ is sparse.

### 2.2 TRANSFORMED COMPONENT ANALYSIS (TCA)

Factor analysis models the distribution over a vector of real-valued sensors $\mathbf{z}$ by the mariginal of a joint distribution over **z** and a smaller number of real-valued latent variables **y**, where

$$p(\mathbf{z},\mathbf{y}) = \mathcal{N}(\mathbf{z};\boldsymbol{\mu} + \boldsymbol{\Lambda}\mathbf{y},\boldsymbol{\Phi})\mathcal{N}(\mathbf{y};\mathbf{0},\mathbf{I}). \quad (5)$$

This model is similar to a probabilistic form of PCA, where the columns of the "factor loading matrix" $\boldsymbol{\Lambda}$ are akin to principal components.

Adding a discrete transformation index to the factor analyzer, we obtain the graphical model shown in Fig. 1d:

$$p(\mathbf{x},\ell,\mathbf{z},\mathbf{y}) = \\ \mathcal{N}(\mathbf{x};\mathbf{G}_\ell\mathbf{z},\boldsymbol{\Psi})\mathcal{N}(\mathbf{z};\boldsymbol{\mu}+\boldsymbol{\Lambda}\mathbf{y},\boldsymbol{\Phi})\mathcal{N}(\mathbf{y};\mathbf{0},\mathbf{I})\rho_\ell, \quad (6)$$

where $\boldsymbol{\mu}$ is the mean of the latent image, $\boldsymbol{\Lambda}$ is a matrix of latent image components (the factor loading matrix) and $\boldsymbol{\Phi}$ is a diagonal noise covariance matrix for the latent image.

Marginalizing over the factors and the latent image gives the transformation conditional likelihood,

$$p(\mathbf{x}|\ell) = \mathcal{N}(\mathbf{x};\mathbf{G}_\ell\boldsymbol{\mu},\mathbf{G}_\ell(\boldsymbol{\Lambda}\boldsymbol{\Lambda}^T+\boldsymbol{\Phi})\mathbf{G}_\ell^T+\boldsymbol{\Psi}), \quad (7)$$

which can be used to compute $p(\mathbf{x})$ and the transformation responsibility $p(\ell|\mathbf{x})$. $\mathbf{G}_\ell(\boldsymbol{\Lambda}\boldsymbol{\Lambda}^T+\boldsymbol{\Phi})\mathbf{G}_\ell^T$ is not sparse, so computing this likelihood exactly takes $N^3$ time. However, the likelihood *can* be computed in linear time if we allow the latent image pixel variances ($\boldsymbol{\Phi}$) to model sensor noise (so, $\boldsymbol{\Psi} = 0$) or assume that the observed noise is smaller than the latent image noise ($|\boldsymbol{\Psi}| << |\boldsymbol{\Phi}|$), in which case $|\mathbf{G}_\ell(\boldsymbol{\Lambda}\boldsymbol{\Lambda}^T+\boldsymbol{\Phi})\mathbf{G}_\ell^T+\boldsymbol{\Psi}| = |\mathbf{G}_\ell(\boldsymbol{\Lambda}\boldsymbol{\Lambda}^T+\boldsymbol{\Phi})\mathbf{G}_\ell^T|$. In our experiments, we used this linear-time computation.

By setting columns of $\boldsymbol{\Lambda}$ equal to the derivatives of $\boldsymbol{\mu}$ with respect to continuous transformation parameters, a TCA can accommodate *both* a local linear approximation and a discrete approximation to the transformation manifold.

### 2.3 MIXTURES OF TRANSFORMED COMPONENT ANALYZERS (MTCA)

A combination of a TMG and a TCA can be used to jointly model clusters, linear components and transformations. Alternatively, a mixture of Gaussians that is invariant to a discrete set of transformations *and* locally linear transformations can be obtained by combining a TMG with a TCA whose components are all set equal to transformation derivatives. The joint distribution $p(\mathbf{x},\ell,\mathbf{z},c,\mathbf{y})$ for the combined model in Fig. 1e is

$$\mathcal{N}(\mathbf{x};\mathbf{G}_\ell\mathbf{z},\boldsymbol{\Psi})\mathcal{N}(\mathbf{z};\boldsymbol{\mu}_c+\boldsymbol{\Lambda}_c\mathbf{y},\boldsymbol{\Phi}_c)\mathcal{N}(\mathbf{y};\mathbf{0},\mathbf{I})\rho_{\ell c}\pi_c.$$

The cluster/transformation likelihood can be computed in linear time as described above for TCA.

## 3 TRANSFORMATIONS IN DYNAMIC BAYES NETS

In Scn. 5, we show that the above static models are useful in a variety of applications that use a set of independent images.

When analyzing video sequences, it is a good idea to make use of temporal coherence, both in the hidden variables that determine the appearances of objects, and in the hidden variables that represent spatial transformations of the objects. We show how the discrete transformation variable described above can be incorporated into the state of a hidden Markov model, producing a "transformed hidden Markov model" (THMM) that efficiently accounts for temporal coherence in class and in spatial transformations. It turns out that if the spatial transformations are highly coherent (*e.g.*, objects in the scene don't jump across the image in one time step), the time needed for inference and learning in the dynamic model is the same as for the equivalent static model applied to the same set of images.

For simplicity, we extend the transformed mixture of Gaussians through time, but the approach described below can also be applied to the transformed mixture of factor analyzers. Fig. 2 shows one possible dynamic Bayes net for this extension, where we assume that the class at time $t$ depends on the class at time $t-1$, but not on the transformation at time $t-1$.

There are several types of independencies that can be incorporated into the THMM. In most cases it is reasonable to assume that while the motion (transformation transition) may or may not depend on the image class, the class transition is independent from the current transformation index: $p(s_t|s_{t-1}) = p(c_t|c_{t-1})p(\ell_t|\ell_{t-1},c_{t-1},c_t)$, where $s_t = (c_t,\ell_t)$ is the combined state. We will also assume the the current transformation does not depend on the current class: $p(s_t|s_{t-1}) = p(c_t|c_{t-1})p(\ell_t|\ell_{t-1},c_{t-1})$.

To model temporal coherence in the transformations, $p(\ell_t|\ell_{t-1},c_{t-1})$ is specified using a map of *relative motion*, $m(\ell_t,\ell_{t-1})$. For example, in the case of image translations, this mapping would simply correspond to the relative shift between two global translations. If a total of $M$ vertical and $M$ horizontal shifts are possible, then the total number of transformations is $L = M^2$ and the transformations can be sorted so that $\ell = iM+j$, where $i,j$ denote indices of the appropriate vertical and horizontal shifts. Then, we can define the mapping as $m(\ell_t,\ell_{t-1}) = \sqrt{(i_t-i_{t-1})^2+(j_t-j_{t-1})^2}$, if we are interested only in the magnitude of the relative motion, or as the vector $m(\ell_t,\ell_{t-1}) = (i_t-i_{t-1},j_t-j_{t-1})$, if the direction of the motion is also important. In a similar fashion the distance measure for other types of transformations can be defined. So, we assume:

$$p(s_t|s_{t-1}) = p(c_t|c_{t-1})p(m(\ell_t, \ell_{t-1})), \text{ or,} \quad (8)$$
$$p(s_t|s_{t-1}) = p(c_t|c_{t-1})p(m(\ell_t, \ell_{t-1})|c_{t-1}), \quad (9)$$

depending on whether or not we want to allow different motion characteristics for different image classes. By assuming small motion between consecutive frames and setting $p(m) = 0$ for $|m| >$ threshold, the number of parameters and the computational load of the inference procedure can be drastically reduced.

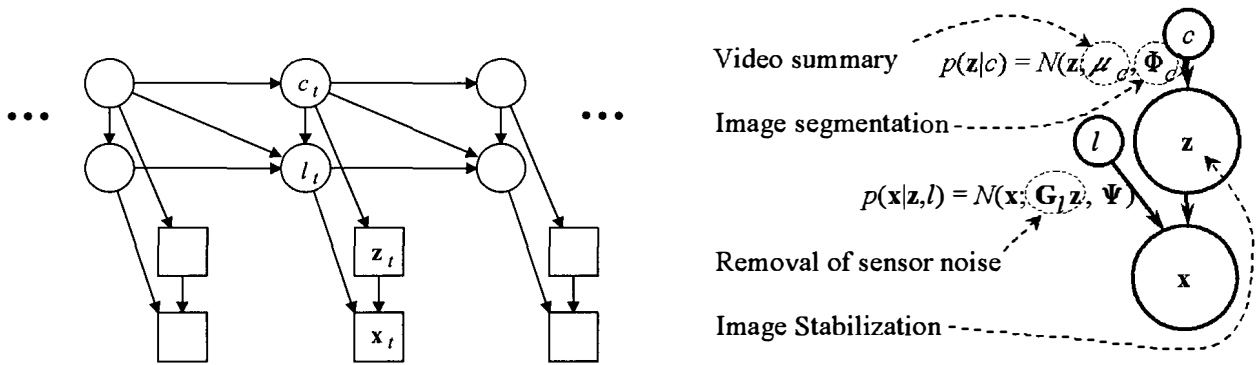

Figure 2: A dynamic Bayes net for transformation-invariant video modeling and a description of the different properties of the video sequence that are extracted when the model is trained using EM.

The THMM has the following parameters: $\mu_c$, for $c = 1, ..., C$ - the mean images for $C$ classes; $\Phi_c$, defining the levels of uncertainty for different pixels for each class; $\Psi$, the diagonal covariance matrix describing the sensor noise; $\pi_s$ where $s = (c, \ell)$, the prior probabilities of different states; and finally the transition probabilities $a_{s,s'} = p(s_t = s'|s_{t-1} = s)$, that can be factorized as shown above. The hidden variables in the model are the states $s_t$ and the latent images $z_t$.

In this generative model, given the previous state, the cluster index $c_t$ and the transformation index $\ell_t$ are drawn randomly from $a_{s_{t-1}, s_t} = p(c_t, \ell_t|c_{t-1}, \ell_{t-1})$. Then, a latent image is drawn from $p(z_t|c_t)$, and then the final frame is drawn from $p(x_t|\ell_t, z_t)$. The process is repeated until the end of the sequence.

If we have the best THMM parameters for a given sequence (estimated using EM as described below), several interesting computer vision tasks can be performed by inference in the THMM (see Fig. 2). Given the input sequence, computing $G_{\ell^{MAP}} \mu_{c^{MAP}}$ removes sensor noise, inferring the transformation index tracks the object and inferring z performs image stabilization. Probabilistic inference is fast, since the state variables of the THMM are discrete. The trained THMM can also be used to classify new video sequences as typical or atypical by applying a threshold to the probability of generating the sequence under the model.

## 4  LEARNING BY EXPECTATION MAXIMIZATION (EM)

In all of the models presented in this paper, the continuous variables are either observed Gaussian variables or are Gaussian variables whose children are Gaussian with linearly related means. So, the hidden continuous variables can be integrated out in closed form using linear algebra, as described in Scn. 2. For the static models, the discrete variables can be lumped together for the purpose of inference and for the dynamic model, the discrete variables at each time step can be lumped together.

Once the models are viewed in this way, the standard expectation maximization (EM) algorithm can be applied in a straightforward fashion to estimate the maximum likelihood parameters (Dempster, et al. 1977). In the E-step, probabilistic inference is used to compute the sufficient statistics. In the M-step, these statistics are used to update the conditional probabilities and densities, while enforcing independence constraints and constraints on the forms of the conditional probabilities.

## 5  EXPERIMENTS

### 5.1  FILTERING SCANNING ELECTRON MICROSCOPE IMAGES

SEM images (e.g., Fig. 3a) can have a very low signal to noise ratio due to a high variance in electron emission rate and modulation of this variance by the imaged material. To reduce noise, multiple images are usually averaged and the pixel variances can be used to estimate certainty in rendered structures. Fig. 3b shows the estimated means and variances of the pixels from 230 SEM images like the ones in Fig. 3a.

We trained a single-cluster TMG with 5 horizontal shifts and 5 vertical shifts on the 230 SEM images using 30 iterations of EM. To keep the number of parameters almost equal to the number of parameters estimated using simple averaging, the transformation probabilities were not learned and the pixel variances in the observed image were set equal after each M step. So, TMG had 1 more parameter than the simple method. Fig. 3c shows the mean and variance learned by the TMG. Compared to simple averaging, the TMG finds sharper, more detailed structure. The variances are significantly lower, indicating that the TMG produces a more confident estimate of the image.

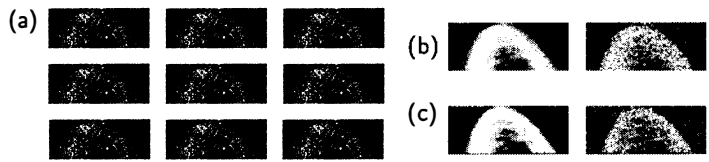

Figure 3: (a) 140 × 56 pixel SEM images. (b) The mean and variance of the image pixels. (c) The mean and variance found by a TMG reveal more structure and less uncertainty.

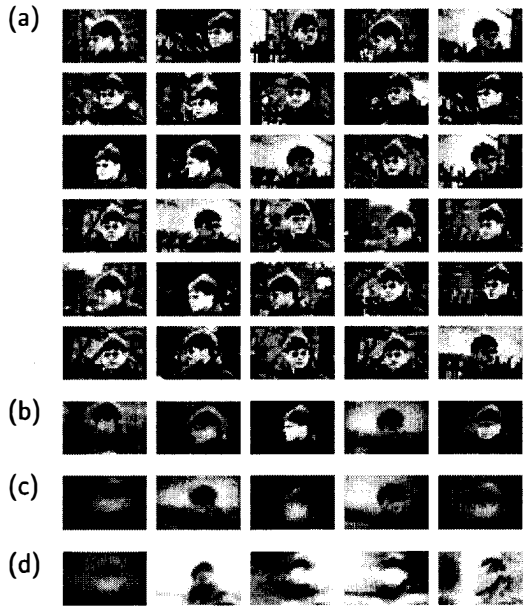

Figure 4: (a) Images of one person with different poses. (b) Cluster means learned by a TMG. (c) Less detailed cluster means learned by a mixture of Gaussians. (d) Mean and first 4 principal components of the data, which mostly model lighting and translation.

### 5.2 CLUSTERING FACIAL POSES

Fig. 4a shows examples from a training set of 400 jerky images of one person with different poses, walking across a cluttered background. We trained a TMG with 5 clusters, 11 horizontal shifts and 11 vertical shifts using 40 iterations of EM. Fig. 4b shows the cluster means, which capture 4 poses and mostly suppress the background clutter. The mean for cluster 4 includes part of the background, but this cluster also has a low mixing proportion of 0.09. A traditional mixture of Gaussians trained using 40 iterations of EM finds blurred means, as shown in Fig. 4c. The first 4 principal components mostly try to account for lighting and translation, as shown in Fig. 4d.

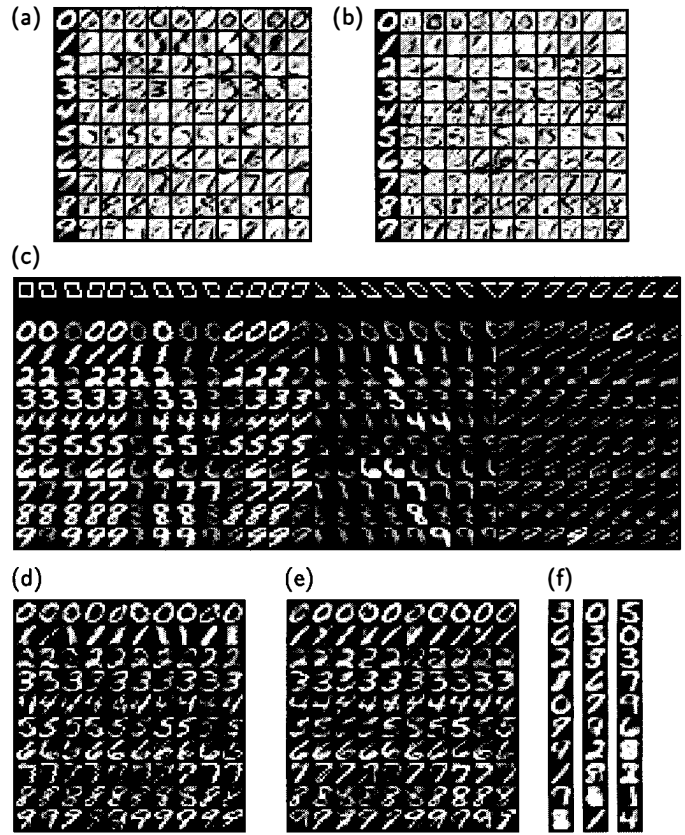

Figure 5: Modeling handwritten digits. (a) Means and components of 10 factor analyzers trained on 200 examples of each digit. (b) Means and components and (c) the sheared + translated means (dimmed transformations have low probability) for each of 10 TCA models trained on the same data. (d) Digits generated from the 10 TCA models and (e) the 10 FA models. (f) The means for a mixture of 10 Gaussians, a mixture of 10 factor analyzers and a 10-cluster TMG trained on all 2000 digits. In each case, the best of 10 experiments was selected using the data likelihood.

### 5.3 MODELING HANDWRITTEN DIGITS

We performed both supervised and unsupervised learning experiments on 8 × 8 greyscale versions of 2000 digits from the CEDAR CDROM (Hull, 1994). Although the preprocessed images fit snugly in the 8 × 8 window, there is wide variation in "writing angle" (e.g., the vertical stroke of the 7 is at different angles). So, we produced a set of 29 shearing+translation transformations (see the top row of Fig. 5c) to use in transformed density models.

In our supervised learning experiments, we trained one 10-component factor analyzer on each class of digit using 30 iterations of EM. The means and components (shown in Fig. 5a) are obfuscated by variation in writing angle (e.g., see the mean for "9"). We also trained one 10-component TCA on each class of digit using 30 iterations of EM. Fig. 5b shows the mean and 10 components for each of the 10 models. The lower 10 rows of images in Fig. 5c show the sheared and translated means. In cases where the transformation probability is below 1%, the image is dimmed. The means found by TCA are sharper and the components tend to account for line thickness and arc size, instead of writing angle. Fig. 5d and e show digits that were randomly generated from the TCAs and the factor analyzers. Since different components in the factor analyzers account for different stroke angles, the simulated digits often have an extra stroke, whereas digits simulated from the TCAs contain fewer spurious strokes.

To test recognition performance, we trained 10-component factor analyzers and TCAs on 200 examples of each digit using 50 iterations of EM. Each set of models used Bayes rule to classify 1000 test patterns and while factor analysis gave an error rate of 3.2%, TCA gave an error rate of only 2.7%.

In our unsupervised learning experiments, we fit 10-cluster mixture models to the entire set of 2000 digits to see which models could identify all 10 digits. We tried a mixture of 10 Gaussians, a mixture of 10 factor analyzers and a 10-cluster TMG. In each case, 10 models were trained using 100 iterations of EM and the model with the highest likelihood was selected and is shown in Fig. 5f. Compared to the TMG, the first two methods found blurred and repeated classes. After identifying each cluster with its most prevalent class of digit, we found that the first two methods had error rates of 53% and 49%, but the TMG had a much lower error rate of 26%.

### 5.4 LEARNING MOTION PATTERNS FROM PAC-MAN MOVIES

Fig. 6 shows some images from a 200 frame long "pac-man movie". Each frame contains one of the four basic 3 × 3 pac-man shapes discernible by the orientation of the mouth. Learning these shapes is actually quite difficult, since the levels of background noise and sensor noise make it difficult to recognize the shapes when the frames are studied individually. However, the pac-man's motion is highly regimented: it always moves by at most one pixel in the direction of his mouth (probability of staying put is 0.2), and occasionally (with probability 0.75) it makes a left turn by appropriately changing the mouth orientation.

A standard clustering method, such as a standard mixture of 6 Gaussians shown in the first row of Fig. 7, will fail to cluster the data properly. Even using a TMG, which treats frames independently, it is difficult to discern the classes, since the shapes differ in less than 2% of the number of observed pixels. For example, in the second row of Fig. 7, we show the 6 classes learned by training a TMG until convergence, while assuming a uniform prior over all $11 \times 11 = 121$ possible translations. While the classes are no longer blurred over the whole image, the background clutter is still not sufficiently suppressed and the shapes are not completely separated. In fact one shape is missing entirely (mouth pointing down).

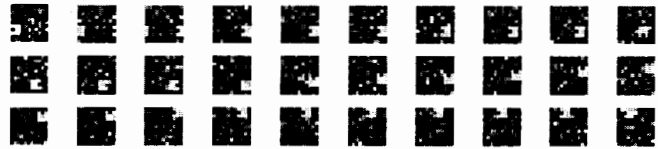

Figure 6: 30 frames from a 200-frame sequence of a pac-man game on an 11x11 grid.

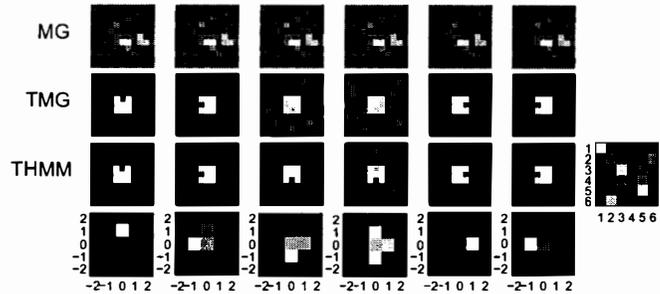

Figure 7: Cluster means resulting form MG, TMG and THMM training; the relative motion patterns for different clusters in the trained THMM (last row) and the class transition matrix of the THMM (right).

One might hope that the extra information available from temporal coherence might make up for the lack of evidence that differentiates the classes. Using a THMM with the transition model in (9), we refined the classes found by TMG by training the THMM using 14 iterations of EM. The background clutter is further suppressed and one of the ambiguous TMG classes was refined into the missing "down shape".

In the fourth row of Fig. 7, we show the motion distribution for each cluster as a gray level image in which the intensity is proportional to the probability of the corresponding relative shift. These images show the dependence between the direction of the mouth in the shape and the direction of the motion.

The class transition probability matrix (shown as a gray level image) captures the left turns in the data as the transition between appropriate appearances. Since 6 classes were used in the model and there are only 4 distinct shapes in the data, 2nd and 6th mean are similar and there is a high probability of switching between them; and the 4th class is estimated to be very unlikely, as the 4 column of the class transition matrix indicates that none of the classes are likely to change into the 4th shape.

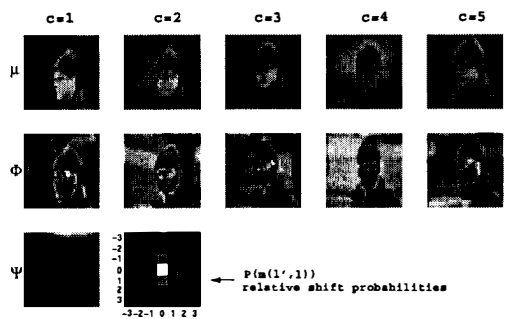

Figure 8: Summarizing the sequence (Movie 1) by the learned THMM parameters.

## 5.5  LEARNING STRUCTURE FROM CLUTTERED, OUTDOOR VIDEOS

The images in Fig. 4a were actually drawn randomly from a 400-frame outdoor video sequence. To learn the motion of the head in the video, we trained a THMM containing $C = 5$ clusters and $L = 625$ translation transformations (25 horizontal shifts and 25 vertical shifts) using 30 iterations of the EM algorithm. We assumed the transition model of (8).

The first row of Fig. 8 shows the resulting cluster means and the second row shows the pixel variances for the clusters (high variance corresponds to high intensity). The last image illustrates the probability distribution of the translational part of the head motion (shifts larger than 3 pixels turned out to be highly unlikely). For or all but one cluster, the learned cluster noise map illustrates that the model learned to segment the object from the background by identifying that the background is much more varying (in cases when the background is uniform, or when it moves together with the object, this would not be the case). The cluster variance map shows how the model scales the contributions of different pixels when matching a class and tracking the object.

To illustrate the various useful inferences shown in Fig. 2, we further degraded the original video sequence by placing a static dark bar in front of both the background and the face (the first row in Fig. 9). We then performed 30 iterations of THMM learning assuming a single class and the same set of transformations as above. On the right of Fig. 9 we show the resulting cluster mean and variances and the sensor noise estimate. The cluster mean and variances show no presence of the black bar and the observation noise $\Psi$ was estimated to be very high in the bar region even though these pixels had almost constant intensities in the image. This happened because the model fixated on the face – the largest object of relatively constant appearance, even though it is partially occluded in many frames. During the training, the missing data was filled in with average appearance of the face in the rest of the sequence (which was possible as the bar did not always occlude the same part of the face). As the face and the background moved with respect to the bar, the THMM could not properly predict the observed pixels in the bar using its single class template and cluster pixel variances, and thus increased the sensor noise in that region.

Removing the sensor noise (second row in Fig. 9), and stabilizing the sequence (third row in Fig. 9) using THMM inference resulted in suppressing the bar – i.e., when inferring the most likely latent image z given the observed frame, the observed pixels in the bar were weighted less than the expected values from the cluster mean, which in the denoised sequence looks as if the black bar was made transparent.

In the final experiment, we set up a mock video street surveillance system, assuming that for the best angle of view the camera had to look at the street through the bars and circular ornaments of a metal gate. A short video sequence (upper row of Fig. 10) shot in this way contained a truck that drove from right to left.

We trained a regular 1-class TMG model (assuming independent frames) and the resulting mean and variances are shown in the left column of Fig. 11. The set of transformations consisted of all possible horizontal shifts in the 60x29 frames, including wrap-around. Even though TMG can normalize for transformations in the data, it learns the static background and the gate, because the foreground bars and circular ornaments occupy too many pixels. In fact, even applying THMM blindly to this problem produces a similar result, since THMM can also determine that there is no global motion in the video.

To overcome this problem, we placed a prior distribution over the relative motion $m(l', l)$ that disallows rest and motion to the right and used it in THMM training. The prior gave equal weights to the motion to the left regardless of the intensity. In this way, we essentially informed the learning algorithm that leftward motion is probable.

The learned mean and variances are shown in the right column of Fig. 11. Even though parts of the truck were occluded in *every* frame, the THMM was able to integrate the truck's appearance over all frames and suppress the gate parts. In the second row of Fig. 10, we demonstrate tracking in this sequence - at each time frame the image contains the cluster mean shifted to the most likely position for that time, given the *whole* observed sequence. As the truck disappears from the screen, the tracking slows down, as the forward-backward estimation of the state sequence forces the motion to continue, but the likelihood of the data drops. Since the transformations wrap around the vertical edges of the frame, the data does not offer evidence for the appropriate shift, since the THMM expects to see the missing part of the truck at the opposite side of the frame.

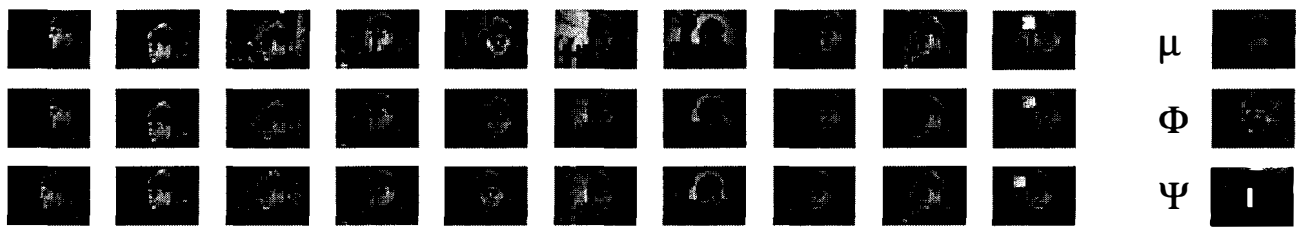

Figure 9: Several frames from a sequence degraded with a black static obstruction occluding the object of interest, and the result of suppressing the distraction (second row) and image stabilization (last row) using a THMM model learned from the degraded sequence. The mean, latent image variances and sensor variances are shown on the right.

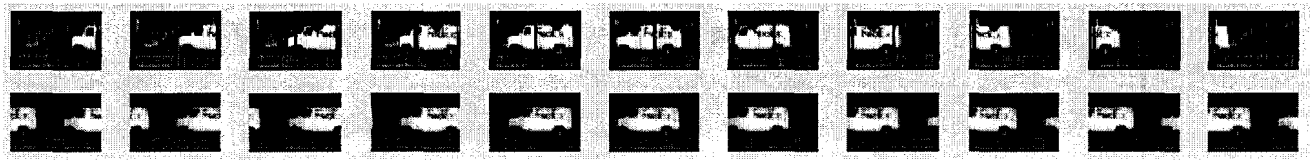

Figure 10: Tracking an object moving between a static background and an obstructing static foreground object (the bar gate).

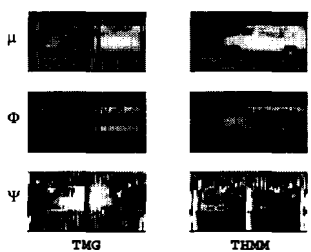

Figure 11: Learned means and variances for a TMG model (left column) and a THMM model (right column) with the relative motion distribution favoring motion to the right.

## 6   SUMMARY

In many learning applications, we know beforehand that the data includes transformations of an easily specified nature (e.g., shearing of digit images) and in the case of time series data, we often know that these transformations are coherent across time. If a generative density model is learned from the data, the model must extract both the transformations and the more interesting and potentially useful structure. We describe a way to add transformation invariance to static and dynamic graphical models, by approximating the transformation manifold with a discrete set of points. This releases the generative model from needing to model the transformations. Although the time needed by this method scales exponentially with the dimensionality of the transformation manifold, we explored a variety of applications where inference and learning were tractable and impressive results were obtained. We are currently exploring faster approximate algorithms, including variational methods and course-to-fine modeling approaches.

## ACKNOWLEDGEMENTS

We used CITO and NSF grants to do this research.